# SpectralCA: Bi-Directional Cross-Attention for Next-Generation UAV Hyperspectral Vision

D.V. Brovko

*Institute of Applied System Analysis, Department of AI*

*National Technical University of Ukraine*

*"Igor Sikorsky Kyiv Polytechnic Institute"*

Kyiv, Ukraine

brovko.dnl@gmail.com ORCID 0009-0008-5676-6845

# ABSTRACT

The work consists of three chapters, includes 12 figures, 4 tables, 31 references, and 1 appendix.

**The relevance of this research** lies in the growing demand for unmanned aerial vehicles (UAVs) capable of operating reliably in complex environments where conventional navigation becomes unreliable due to interference, poor visibility, or camouflage. Hyperspectral imaging (HSI) provides unique opportunities for UAV-based computer vision by enabling fine-grained material recognition and object differentiation, which are critical for navigation, surveillance, agriculture, and environmental monitoring.

**The aim of this work** is to develop a deep learning architecture integrating HSI into UAV perception for navigation, object detection, and terrain classification. Objectives include: reviewing existing HSI methods, designing a hybrid 2D/3D convolutional architecture with spectral-spatial cross-attention, training, and benchmarking.

**The methodology** is based on the modification of the Mobile 3D Vision Transformer (MDvT) by introducing the proposed SpectralCA block. This block employs bi-directional cross-attention to fuse spectral and spatial features, enhancing accuracy while reducing parameters and inference time. Experimental evaluation was conducted on the WHU-Hi-HongHu dataset, with results assessed using Overall Accuracy, Average Accuracy, and the Kappa coefficient.

**The findings confirm** that the proposed architecture improves UAV perception efficiency, enabling real-time operation for navigation, object recognition, and environmental monitoring tasks.

A version of this work has been accepted for presentation at the *2025 IEEE 8th International Conference on Methods and Systems of Navigation and Motion Control*.

# CONTENTS

# INTRODUCTION

Unmanned aerial vehicles (UAVs) have become an essential platform for a wide spectrum of applications, ranging from environmental monitoring and precision agriculture to security, mapping, and autonomous navigation. Their onboard perception systems typically rely on GPS, inertial measurement units, and RGB cameras. However, these modalities face limitations when external conditions such as signal loss, fog, camouflage, or spectral similarity reduce their effectiveness.

Hyperspectral imaging (HSI) offers a rich source of information by capturing reflectance across hundreds of spectral bands, enabling UAVs to differentiate between materials and objects that appear indistinguishable in standard RGB images. This capability makes HSI particularly valuable not only for navigation in degraded environments, but also for object detection, land-cover classification, and fine-grained environmental analysis. For example, UAVs equipped with HSI can recognize vegetation stress, detect man-made structures, and classify terrain types with far greater precision than conventional systems.

The proposed SpectralCA framework integrates hyperspectral sensing with deep learning, specifically a cross-attention mechanism that fuses spectral and spatial features. The objective of this research is to investigate how such integration can enhance UAV autonomy by simultaneously improving navigation accuracy and computer vision capabilities such as object recognition, mapping, and environmental perception.

# CHAPTER 1 THEORETICAL INFORMATION

## 1.1 Hyperspectral Images: Definition and Application

A hyperspectral image (HSI) is a three-dimensional data array, or hypercube, capturing a continuous reflection spectrum for each pixel across dozens or hundreds of narrow wavelength bands. HSI combines digital photography with spectroscopy, gathering information across the electromagnetic spectrum. Unlike standard RGB images with only 3 channels, hyperspectral images contain dozens or hundreds of spectral bands with very high spectral resolution. This detailed pixel-level spectrum enables identification of spectral signatures – unique material "fingerprints" used to recognize objects and their composition [1]. For example, characteristic spectra can identify chlorophyll in plants or petroleum hydrocarbons in soil.

Hyperspectral imaging technology originated in Earth remote sensing. NASA began developing hyperspectral scanners for aerospace applications in the 1980s [3]. Over three decades, HSI has become a powerful tool providing phenomenal spectral and spatial detail for Earth surface studies [8].

Fig. 1.1 shows the WHU-Hi-HongHu dataset, one of the most common HSI datasets.

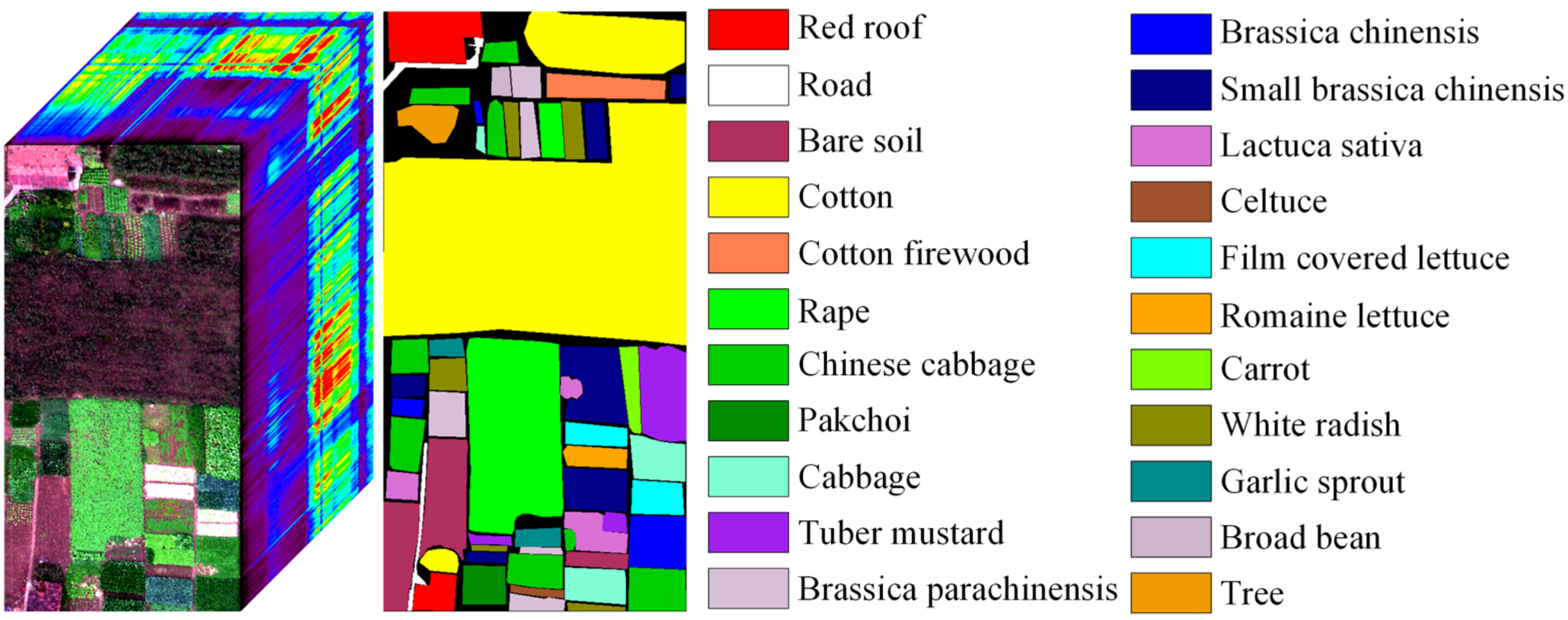

**Fig. 1.1** – WHU-Hi-HongHu dataset [28]

Initially used for mineral mapping, agriculture, and land-use planning, the technology has expanded to other fields. HSI's information richness enables diverse military, environmental, and civilian applications. In remote sensing, HSI is used for detailed land cover classification, vegetation typing, soil and water monitoring, and detecting camouflage or mines. In agriculture, hyperspectral imaging assesses crop health, identifies plant species, and detects pest or disease damage [8]. For food security and industry, HSI enables non-invasive quality checks and contamination detection. Medical applications are equally promising: tissue imaging provides diagnostic information about physiology and morphology without biopsy [2]. For example, HSI data can distinguish healthy from cancerous tissues by spectral features. Hyperspectral systems are thus considered promising for tasks ranging from environmental monitoring to quality control and disease diagnosis.

This versatility has made HSI a focus of intensive analysis. A key task is HSI classification – automatically recognizing each pixel's class based on the material or object it represents.

## 1.2 Methods for Classifying Hyperspectral Images

### 1.2.1 CNN

Convolutional Neural Networks (CNNs) are widely used for hyperspectral image (HSI) classification due to their ability to automatically extract spatial and spectral features. Different variants exist: 1D-CNNs process spectral vectors, 2D-CNNs capture spatial context, and 3D-CNNs jointly model spatial–spectral interactions. Hybrid designs such as HybridSN combine 3D and 2D convolutions (Fig. 1.2), improving classification accuracy by capturing both local textures and spectral signatures. However, 3D-CNNs are computationally expensive, with a large number of parameters and high training time, which can limit their deployment on UAVs.

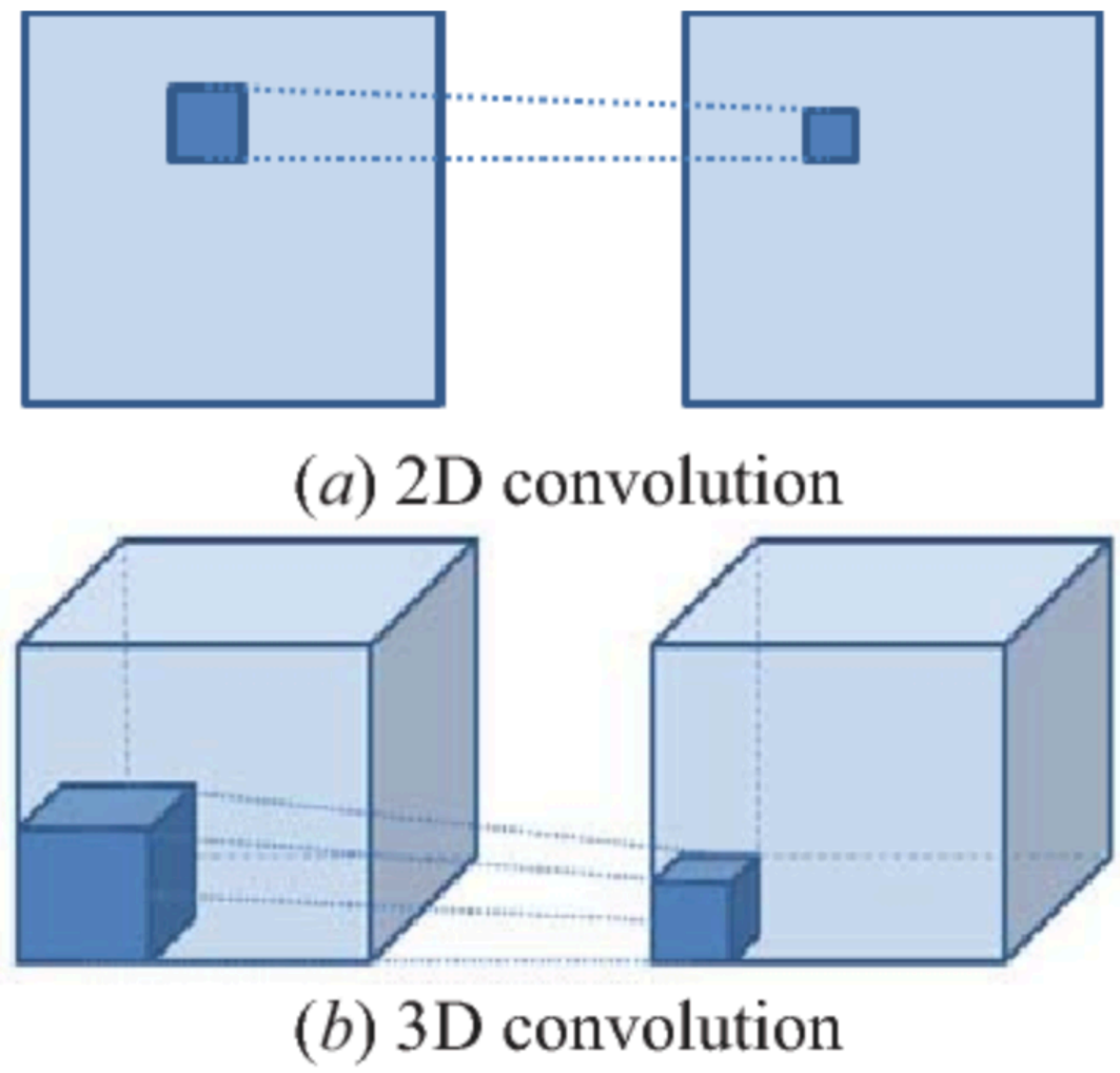

**Fig. 1.2** – Comparison of (a) 2D convolution and (b) 3D convolution [4]

### 1.2.2 Transformers

Transformers, originally developed for natural language processing, have been adapted to HSI analysis through the self-attention mechanism, which models long-range dependencies across spectral and spatial domains. Unlike CNNs, which focus on local receptive fields, Transformers can capture global relationships, enabling more precise material and terrain classification. Models such as SpectralFormer (Fig. 1.3) treat the spectral bands of a pixel as sequences, achieving state-of-the-art accuracy in HSI classification.

In UAV remote sensing, fusing hyperspectral imagery with LiDAR data provides complementary information: spectral features describe material properties, while LiDAR offers structural and elevation data. Recent methods employ cross-attention Transformers, where LiDAR tokens act as queries and HSI features as keys/values (Fig. 1.4). This allows the model to align elevation and spectral features, enhancing land-cover classification and terrain analysis. For UAV navigation, such fusion enables more robust obstacle detection, terrain mapping, and environmental monitoring in complex environments.

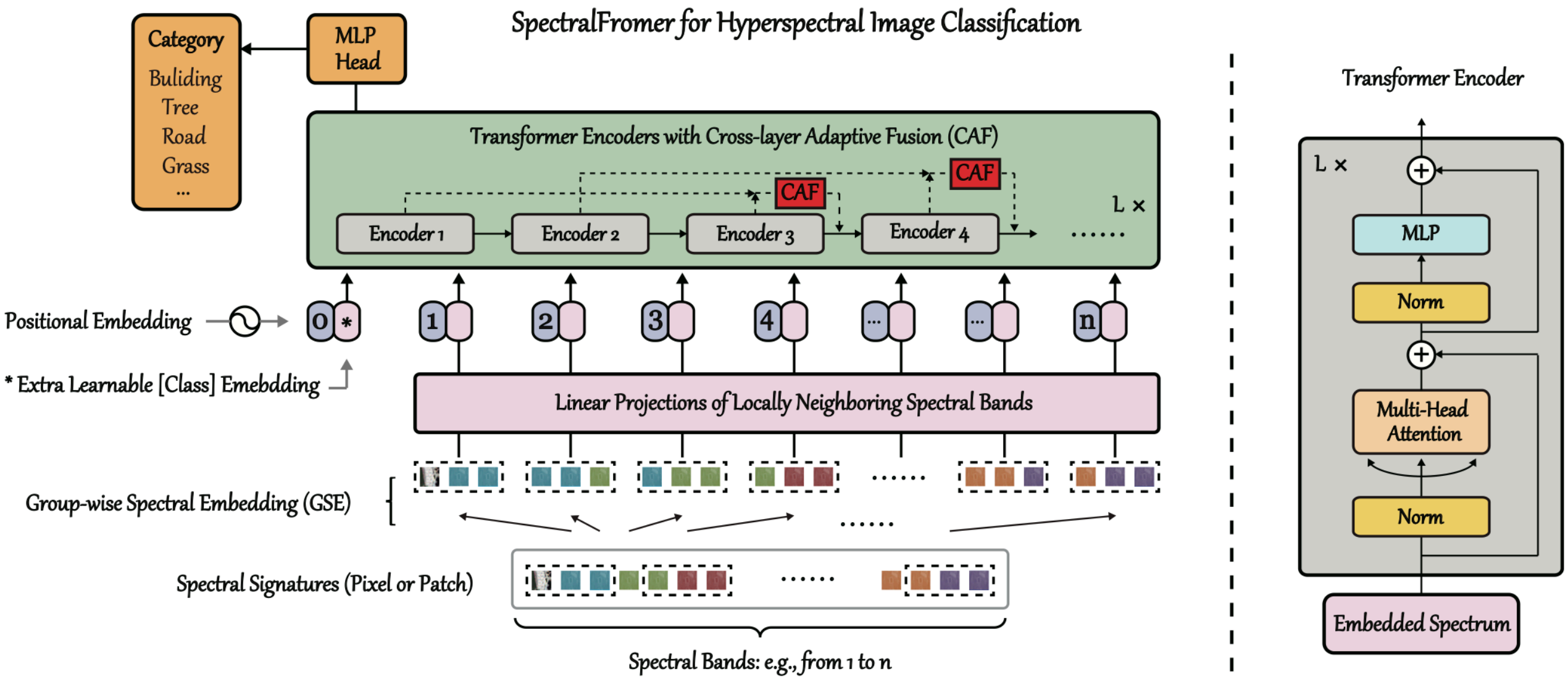

**Fig. 1.3** – Topology of the SpectralFormer model [16]

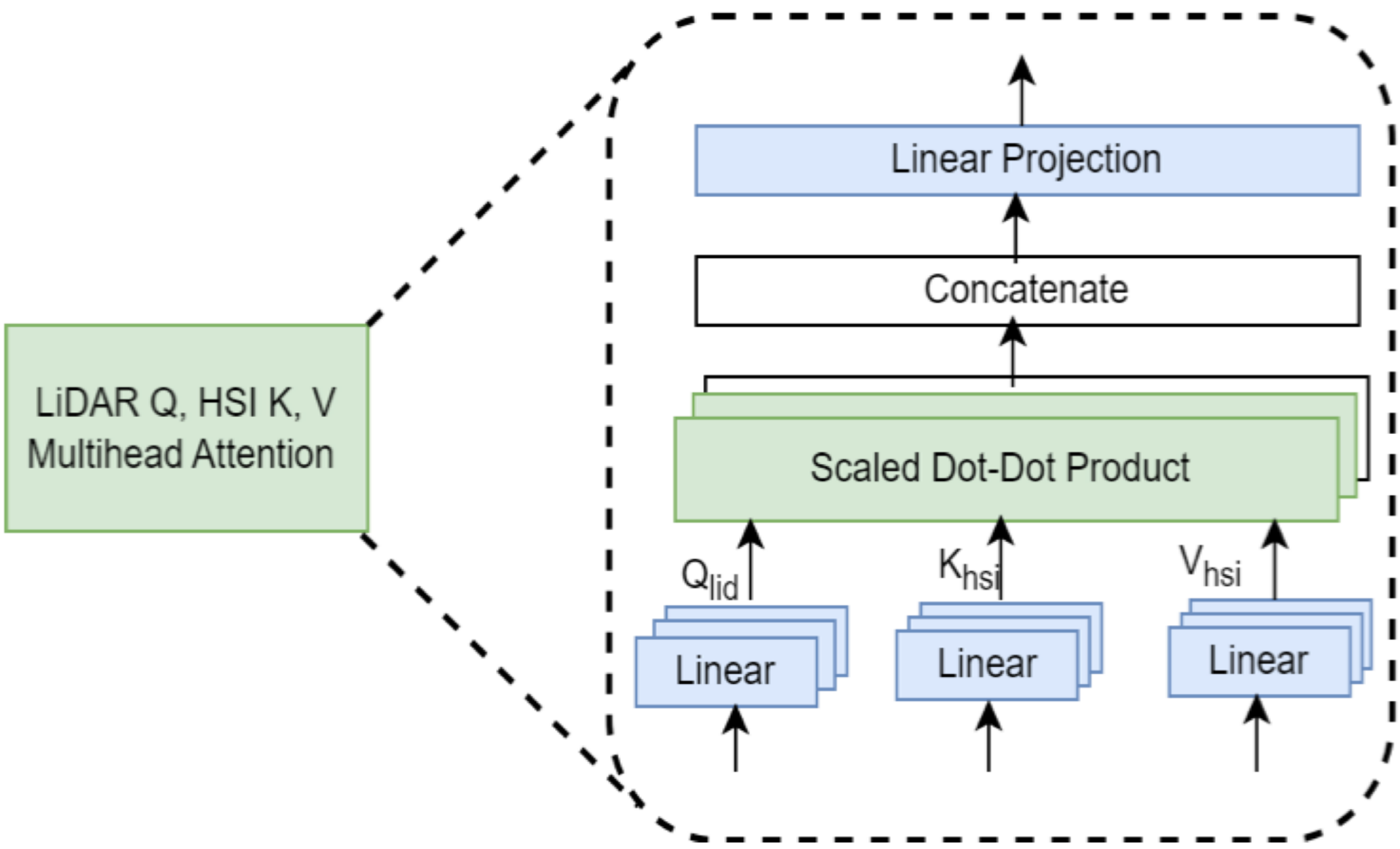

**Fig. 1.4** – Cross-attention combining LiDAR and HSI [18]

### 1.2.3 Hybrid Networks

The Mobile 3D Convolutional Vision Transformer (MDvT) integrates lightweight 3D convolutions with Vision Transformer blocks (Fig. 1.5). The 3D convolutions extract local spectral–spatial features, while the Transformer captures global dependencies. By embedding mobile convolutional layers (inspired by MobileNet inverted residual blocks), MDvT reduces parameter count and accelerates inference, making it suitable for UAV onboard processing. This hybrid approach achieves higher accuracy than pure CNN or Transformer architectures while maintaining efficiency.

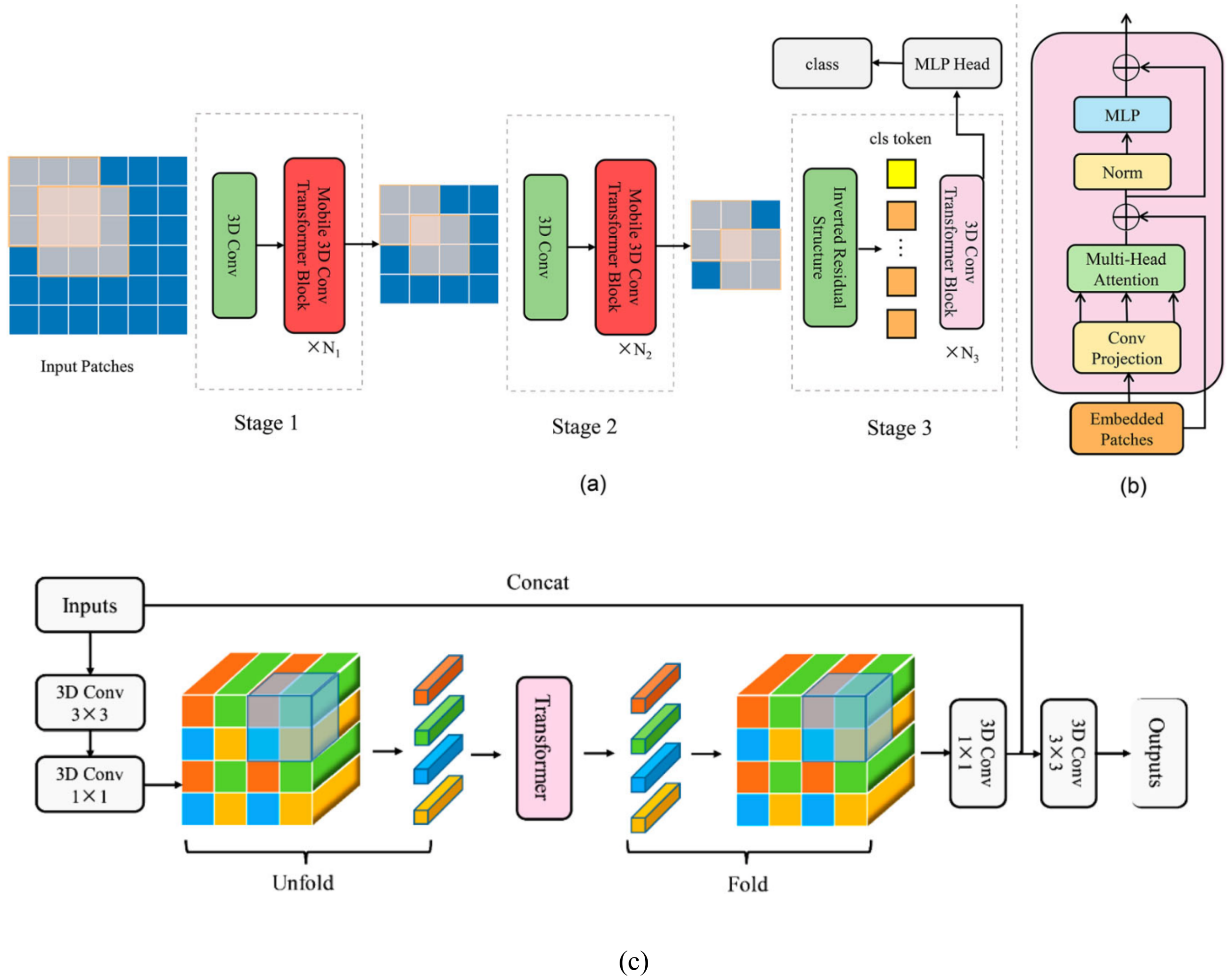

**Fig. 1.5** – (a) overall topology of the MDvT model;
(b) 3D Conv Transformer Block; (c) Mobile 3D Conv Transformer Block [19]

### 1.3 Semi-Supervised Learning

Deep learning demonstrates high accuracy in classifying hyperspectral images (HSI), but its effectiveness largely depends on the amount of labeled data available. In HSI tasks, labeling pixels for training is extremely time-consuming and resource-intensive, as each pixel is associated with a spectrum and often requires consultation with experts in geology, agronomy, or military affairs [22]. This results in a limited number of labeled examples, especially in new or critical regions such as combat zones or contaminated areas.

In such conditions, semi-supervised learning (SSL) is a promising direction, allowing both labeled and unlabeled samples to be used during model training. The main idea behind SSL is that, despite the small number of annotated examples, the model can learn the general structure of data distribution using a large amount of unlabeled examples. The main approaches to SSL in HSI are:

1. Self-learning (the model first learns on a small labeled sample, and then uses its predictions on unlabeled data as pseudo-labels for further training) [6].
2. Consistent learning (the idea is that the model should predict the same results for the same example, even if random augmentations are applied to it) [7].
3. Pseudo-labelling (the model receives pseudo-labels on unlabelled data, but includes only those samples in further training for which it is most confident) [30].
4. Graph models (the use of graph neural networks (GNN) in combination with semi-supervised learning allows information to be transferred between pixels that are spatially or spectrally close) [31].

For Ukraine, where monitoring agricultural land, detecting mines, pollution zones, etc. is extremely important, semi-supervised models pave the way for creating effective systems based on a limited amount of verified data. Instead of manually annotating thousands of pixels, you can use a small amount of labeled data combined with a large number of unlabeled images obtained from drones or satellites. This significantly reduces costs and speeds up the process of implementing HSI analysis in practice.

## 1.4 Problem Statement

To achieve high-performance UAV computer vision and navigation using HSI, the following tasks must be addressed:

1. conduct a review of existing approaches to hyperspectral image classification;
2. design a hybrid model architecture based on MDvT and Cross-Attention Fusion;
3. implement dual-branch processing of spectral and spatial features followed by their fusion;
4. perform training and evaluation of the model on standard hyperspectral datasets;
5. carry out a comparative analysis with classical methods (SVM, 3D-CNN, Vision Transformer);
6. evaluate the model's accuracy, inference time, and number of parameters.

The multi-objective optimization formula, which simultaneously minimizes classification error, inference time, and the number of model parameters, is defined as:

$$\min_{\theta}[\lambda_1 \cdot \mathrm{E}(\theta) + \lambda_2 \cdot \frac{\mathrm{T}\ (\theta)}{\mathrm{T_{ref}}} + \lambda_3 \cdot \frac{\mathrm{P}(\theta)}{\mathrm{P_{ref}}}]$$

where $\theta$ – vector of model parameters (weights, biases),

$\mathrm{E}(\theta) = 1 - \mathrm{A}(\theta)$ – classification error,

$\mathrm{T}(\theta)$ – average inference time in seconds,

$\mathrm{P}(\theta)$ – number of model parameters in millions,

$\lambda_1, \lambda_2, \lambda_3 > 0$, $\lambda_1 + \lambda_2 + \lambda_3 = 1$ – weight coefficients defining the importance of each criterion,

$\mathrm{T_{ref}}$, $\mathrm{P_{ref}}$ - reference values of inference time and parameters of the baseline model.

In this work, the main priority is optimization of inference time and model size, in order to make the architecture applicable on mobile devices under dynamic real-time conditions.

For classification quality evaluation, the following metrics will be used:

- Overall Accuracy (OA):

$$\mathrm{OA} = \frac{1}{M} \sum_{k=1}^{M} \mathbf{1}(y_k = \hat{y}_k)$$

- Average Accuracy (AA):

$$\mathrm{AA} = \frac{1}{C} \sum_{c=1}^{C} \frac{1}{|M_c|} \sum_{k \in M_c} \mathbf{1}(y_k = \hat{y}_k)$$

- Kappa Coefficient:

$$\varkappa = \frac{p_o - p_e}{1 - p_e}$$

- as well as average inference time and number of model parameters.

# CHAPTER 2 INTRODUCING SPECTRAL CROSS-ATTENTION

This section presents the design of a modified MDvT architecture that replaces the basic MobileViTBlock unit with a new SpectralCA (Spectral Cross-Attention) module. The source code is in Appendix A.

## 2.1 SpectralCA Block Topology

The research question addressed in this work is how hyperspectral data can be effectively incorporated into UAV navigation, object detection, material recognition, and terrain classification pipelines. To answer this, the SpectralCA block was designed as a hybrid module that combines 2D and 3D convolutions with bidirectional cross-attention. This section presents the construction of a modified MDvT architecture, in which the baseline MobileViTBlock is replaced with a new module – SpectralCA (Spectral Cross-Attention) (Fig. 2.1).

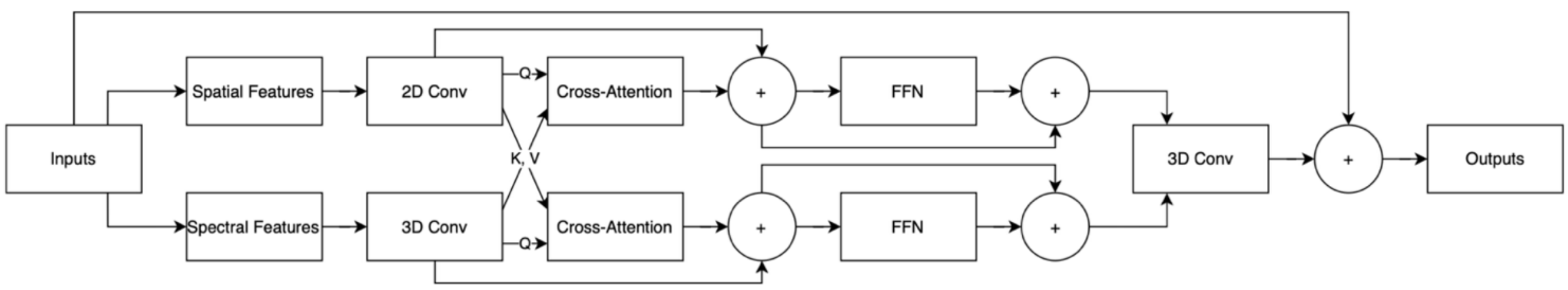

**Fig. 2.1** – Topology of the SpectralCA block

The purpose of the SpectralCA block is to improve the efficiency of hyperspectral image classification by enhancing the interaction between spatial and spectral features, while simultaneously reducing the number of parameters and inference time.

## 2.2 Mathematical Model

The input to the module is a hyperspectral image tensor:

$$X \in R^{\{B \times C \times H \times W \times D\}},$$

where:

$B$ – batch size, $C$ – number of channels, $H \times W$ – spatial coordinates, $D$ – spectral depth.

In this study, we employ hyperspectral data from the WHU-Hi-HongHu agricultural dataset, which contains high-precision spectral measurements of farmland [28].

The SpectralCA architecture processes features through two parallel paths, followed by a cross-attention mechanism:

Spatial Feature Extraction. The mean value across the spectral axis produces a 2D spatial feature map:

$$X_{spatial} \in \mathbb{R}^{B \times C \times H \times W},$$

$$F_{spatial} = \mathrm{SiLU}(\mathrm{BN}(\mathrm{Conv2D}^{3\times3}(X_{spatial}))),$$

where Conv2D detects local textures, BatchNorm stabilizes training, and SiLU ensures nonlinear activation without vanishing gradients. Layer normalization is then applied:

$$S = \mathrm{LayerNorm}(F_{spatial}).$$

Spectral Feature Extraction. A 3D convolution with a 3×3×3 kernel jointly models spatial–spectral correlations:

$$F_{spectral} = \mathrm{SiLU}(\mathrm{BN}(\mathrm{Conv3D}^{3\times3\times3}(X))),$$

followed by:

$$P = \mathrm{LayerNorm}(F_{spectral}).$$

Cross-Attention Mechanism. Two attention branches are introduced:

$$Att_1 = \text{softmax}\left(\frac{Q_s K_p^T}{\sqrt{d_h}}\right) V_p, \quad Att_2 = \text{softmax}\left(\frac{Q_p K_s^T}{\sqrt{d_h}}\right) V_s,$$

where $Q$, $K$, $V$ are query, key, and value matrices; $d_h=d/h$ with $d$ being feature dimensionality and $h$ the number of attention heads.

$Att_1$ allows spatial context to strengthen relevant spectral components (e.g., crop type recognition).

$Att_2$ enables spectral vectors to adapt to positional context, improving spatial adaptability.

Residual connections, normalization, and feed-forward networks (FFNs) are applied in both branches:

$$S' = S + \text{Dropout}(Att_1), \quad S'' = S' + \text{FFN}(\text{LayerNorm}(S')),$$

$$P' = P + \text{Dropout}(Att_2), \quad P'' = P' + \text{FFN}(\text{LayerNorm}(P')).$$

Finally, results are concatenated and merged with a 3D projection, followed by a global residual connection:

$$Z = \text{Concat}(S, P), \quad Y = \text{Conv3D}^{1\times1\times1}(Z) + X.$$

This ensures preservation of initial information while enhancing gradient flow and reducing overfitting.

### 2.3 Training Parameters

The distribution of training parameters for the MDvT architecture with the new SpectralCA block is shown in Fig. 2.2.

For the first SpectralCA block (configuration: in_channels=32, transformer_dim=64), the module contains 383,680 parameters. In an extended configuration (in_channels=64, transformer_dim=96), the parameter count increases to 816,568.

```
==========================================================================================
Layer (type:depth-idx)                                        Param #
==========================================================================================
MDvT                                                          256
├─Sequential: 1-1                                             --
│    └─Conv3d: 2-1                                            1,792
│    └─BatchNorm3d: 2-2                                       128
│    └─ReLU: 2-3                                              --
├─Sequential: 1-2                                             --
│    └─SpectralCA: 2-4                                        --
│    │    └─Sequential: 3-1                                   55,584
│    │    └─Sequential: 3-2                                   166,176
│    │    └─CrossAttention: 3-3                               74,496
│    │    └─LayerNorm: 3-4                                    192
│    │    └─LayerNorm: 3-5                                    192
│    │    └─LayerNorm: 3-6                                    192
│    │    └─LayerNorm: 3-7                                    192
│    │    └─Sequential: 3-8                                   37,152
│    │    └─Sequential: 3-9                                   37,152
│    │    └─Conv3d: 3-10                                      12,352
├─Sequential: 1-3                                             --
│    └─Conv3d: 2-5                                            221,312
│    └─BatchNorm3d: 2-6                                       256
│    └─ReLU: 2-7                                              --
├─Sequential: 1-4                                             --
│    └─SpectralCA: 2-8                                        --
│    │    └─Sequential: 3-11                                  138,600
│    │    └─Sequential: 3-12                                  415,080
│    │    └─CrossAttention: 3-13                              116,160
│    │    └─LayerNorm: 3-14                                   240
│    │    └─LayerNorm: 3-15                                   240
│    │    └─LayerNorm: 3-16                                   240
│    │    └─LayerNorm: 3-17                                   240
│    │    └─Sequential: 3-18                                  57,960
│    │    └─Sequential: 3-19                                  57,960
│    │    └─Conv3d: 3-20                                      30,848
├─InvertedResidual: 1-5                                       --
│    └─Sequential: 2-9                                        --
│    │    └─ConvBNReLU: 3-21                                  99,840
│    │    └─ConvBNReLU: 3-22                                  22,272
│    │    └─Conv3d: 3-23                                      196,608
│    │    └─BatchNorm3d: 3-24                                 512
├─Sequential: 1-6                                             --
│    └─Rearrange: 2-10                                        --
│    └─LayerNorm: 2-11                                        512
├─Sequential: 1-7                                             --
│    └─Transformer: 2-12                                      --
│    │    └─ModuleList: 3-25                                  4,876,800
├─Dropout: 1-8                                                --
├─Sequential: 1-9                                             --
│    └─LayerNorm: 2-13                                        512
│    └─Linear: 2-14                                           5,654
==========================================================================================
Total params: 6,627,702
Trainable params: 6,627,702
Non-trainable params: 0
==========================================================================================
```

**Fig. 2.2** – Distribution of MDvT training parameters with SpectralCA

A finalized detailed description of SpectralCA components is provided in the Table 2.3.

**Table 2.3** – Parameter distribution of SpectralCA components in the modified MDvT architecture

| Component of SpectralCA | Description | Parameter count (in_channels=32, transformer_dim=64) | Parameter count (in_channels=64, transformer_dim=96) |
|---|---|---|---|
| **Spatial Conv Block** | Conv2D + Batch Normalization + SiLU, 3×3 kernel | 55 584 | 138 600 |
| **Spectral Conv Block** | Conv3D + Batch Normalization + SiLU, 3×3×3 kernel | 166 176 | 415 080 |
| **Cross-Attention (2 directions)** | 6×Linear + 2×Linear out | 74 496 | 116 160 |
| **LayerNorm ×4** | Normalization for both paths | 768 | 960 |
| **FFN for Spatial features** | 2×Linear, SiLU, Dropout | 37 152 | 57 960 |
| **FFN for Spectral features** | 2×Linear, SiLU, Dropout | 37 152 | 57 960 |
| **Output Projector** | Conv3D (1×1×1) for channel restoration | 12 352 | 30 848 |
| **Total** | | 383 680 | 816 568 |

The SpectralCA block can be integrated into MDvT either as a replacement for the MobileViTBlock (the approach described in the work) or as an additional component in the early layers. Its design allows adaptive weighting of spectral components with respect to spatial context, while maintaining the geometric structure of hyperspectral

imagery – a crucial property for UAV visual perception tasks such as terrain recognition, obstacle detection, and environmental monitoring.

# CHAPTER 3 APPLICATION

## 3.1 Comparative Analysis of SpectralCA and MobileViTBlock as Part of MDvT

In the base MDvT model, the MobileViTBlock was replaced with the proposed SpectralCA block. The training was performed on the WHU-Hi-HongHu dataset, used for object detection tasks. The progress is shown in Fig. 3.1.

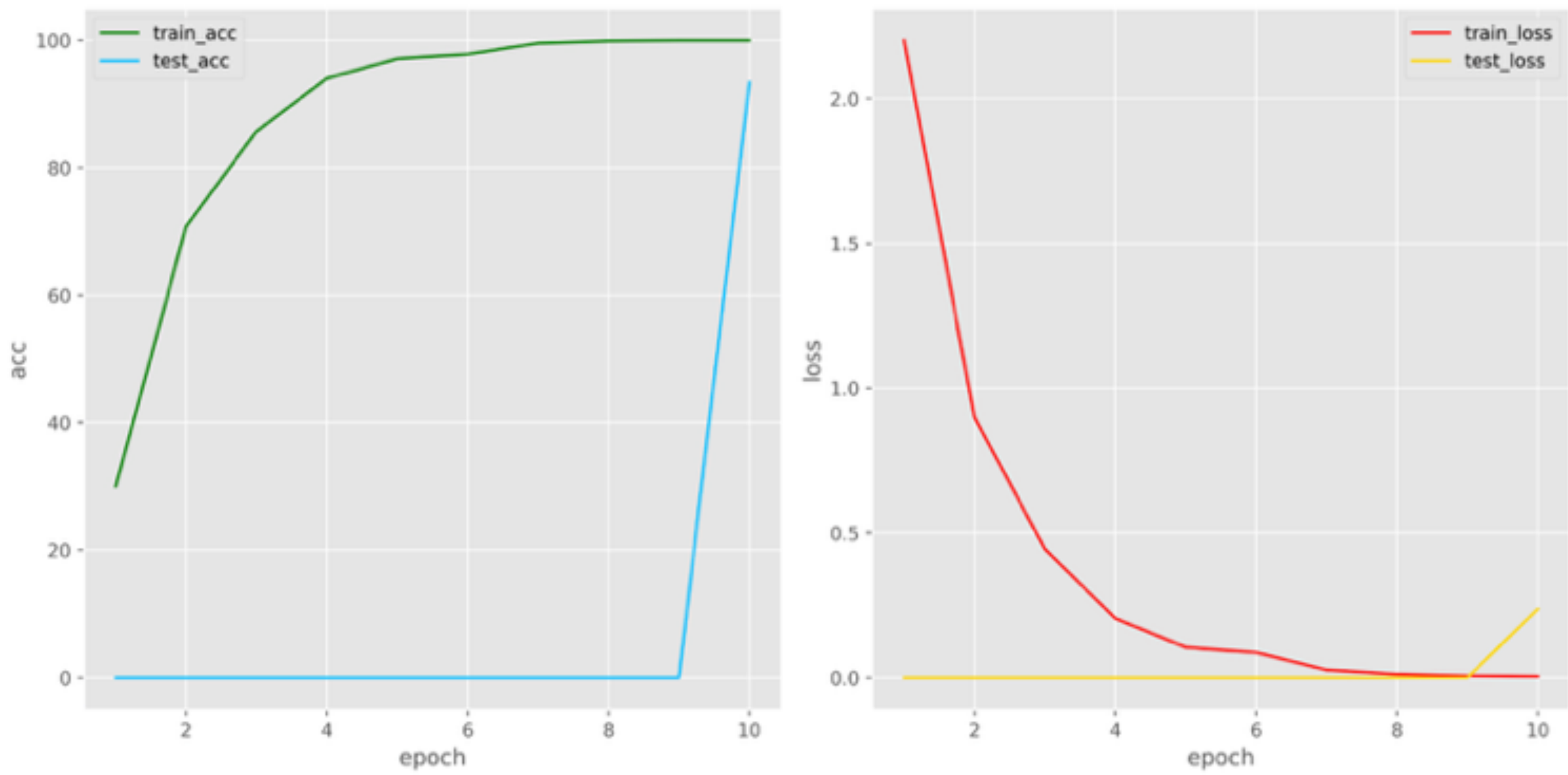

**Fig. 3.1** – Training results using the SpectralCA block: accuracy (left) and loss (right)

The analysis shows that the modified MDvT architecture has approximately 1.1 million fewer parameters compared to the original version (Fig. 3.2).

Experiments demonstrate that the SpectralCA architecture is almost twice as fast as the base MDvT model, while maintaining only a 4% drop in accuracy, reaching approximately 93%. This performance boost is especially important for real-time inference. The comparison is illustrated in Fig. 3.3, with a more detailed report provided in Table 3.4.

For UAV perception, this translates into more reliable detection of terrain classes, vegetation stress, and artificial structures. Improved perception capabilities directly enhance UAV motion control by reducing navigation errors and enabling safer path planning.

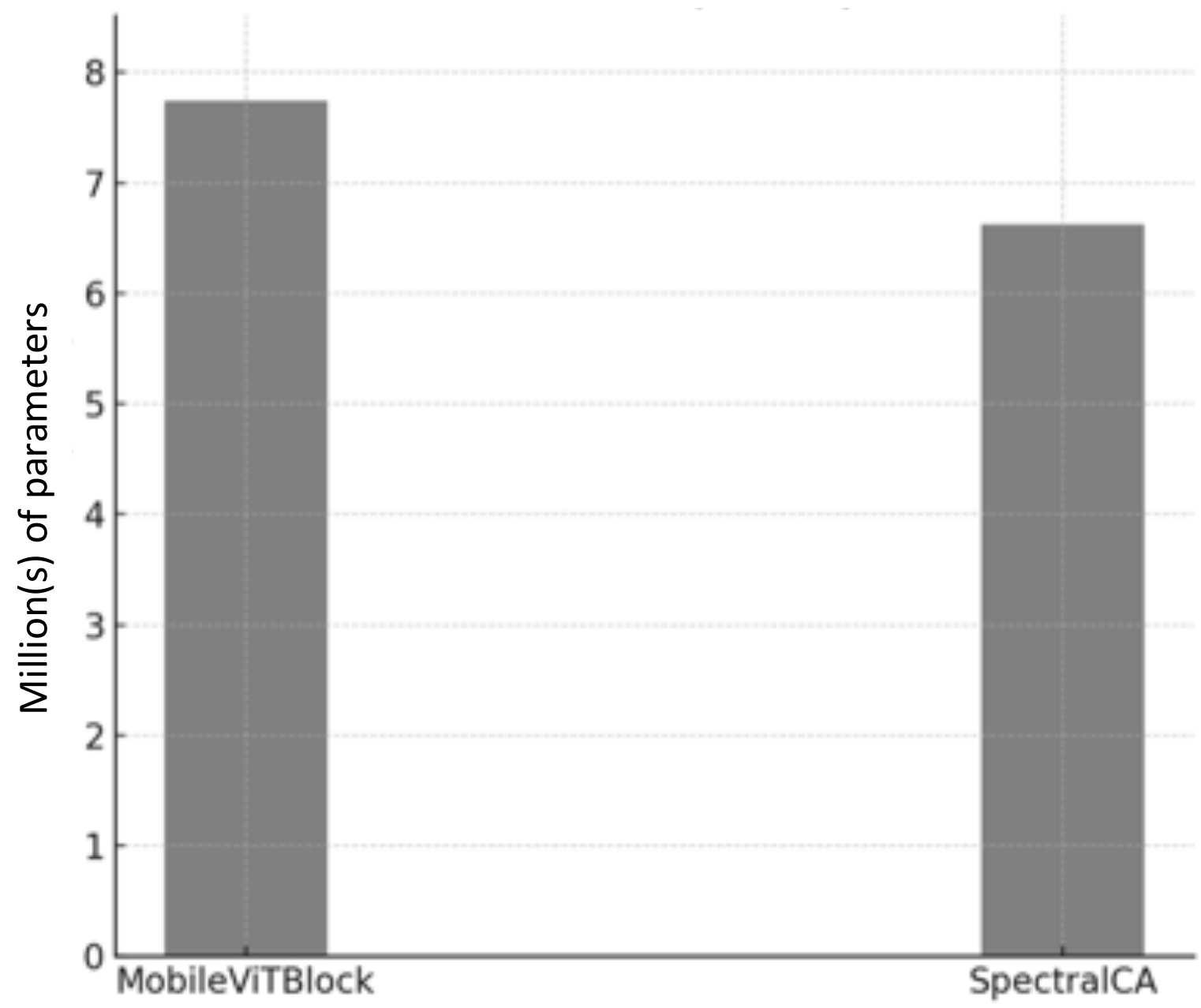

**Fig. 3.2** – Comparison of the model parameter counts (in millions) between the SpectralCA and MobileViTBlock variants

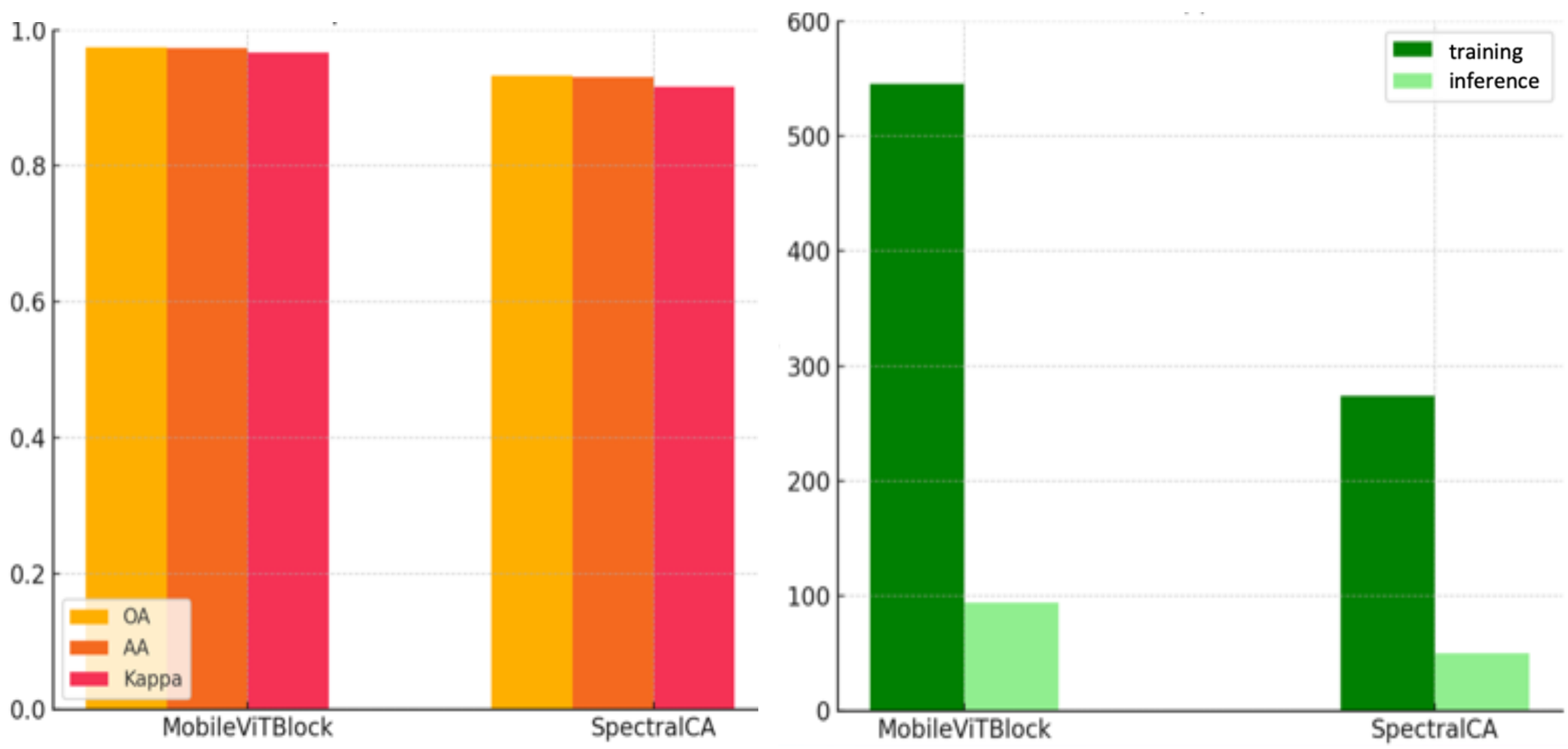

**Fig. 3.3** – Comparison between the proposed SpectralCA block and the base MobileViTBlock: accuracy (left) and training/inference time in seconds (right)

**Table 3.4** – Detailed report on the performance of the SpectralCA and MobileViTBlock variants

| Model | OA | AA | Kappa | Train time (sec) | Infer time (sec) | Param count (mil) |
|---|---|---|---|---|---|---|
| **MobileViTBlock** | **0.974** | **0.9739** | **0.9671** | 545.64 | 94 | 7.746 |
| **SpectralCA** | 0.9334 | 0.9308 | 0.9162 | **274.38** | **50** | **6.628** |

Furthermore, robustness tests show that SpectralCA maintains performance under variable illumination and partial occlusion.

Key advantages of using SpectralCA compared to MobileViTBlock:

- x2 training and inference speed,
- 1.1 million fewer parameters in the final MDvT model,
- accuracy > 93% preserved.

These findings confirm that integrating hyperspectral imaging with cross-attention deep learning provides a practical pathway to advancing UAV autonomy.

## 3.2 Semi-Supervised Learning

The task of processing hyperspectral images using AI is often accompanied by a serious limitation – a lack of labeled data. Each pixel of a hyperspectral cube corresponds to a spectral vector, which in many cases requires expert interpretation, especially when it comes to specific fields such as agronomy, geology, or military intelligence. Therefore, labeling hyperspectral data is a labor-intensive and expensive process. At the same time, modern neural networks, especially transformer and hybrid architectures, have a large number of parameters and require a significant amount of data for effective training. This creates a contradiction between the need for a large number of labeled samples and the practical impossibility of obtaining them in real conditions.

One effective solution to this problem is semi-supervised learning, which allows a small number of annotated samples to be combined with a large amount of unlabeled data. Within the scope of this work, a self-training approach using proxy labeling was applied. This method assumes that the model is first trained on a limited number of manually labeled pixels and then used to generate pseudo-labels for unlabeled samples. The model's most confident predictions are included in a new training set, after which the model is retrained on the expanded dataset. The procedure is performed iteratively, allowing for a gradual increase in the amount of effectively used data.

In the proposed implementation, a heuristic confidence threshold was used: only those pseudo-labels for which the classification probability exceeded a specified level (namely 90%) were included in further training. This avoids the accumulation of noise in the training set and improves the overall stability of the model.

The results of the model trained on a dataset with pseudo-labels are shown in Fig. 3.5 and 3.6.

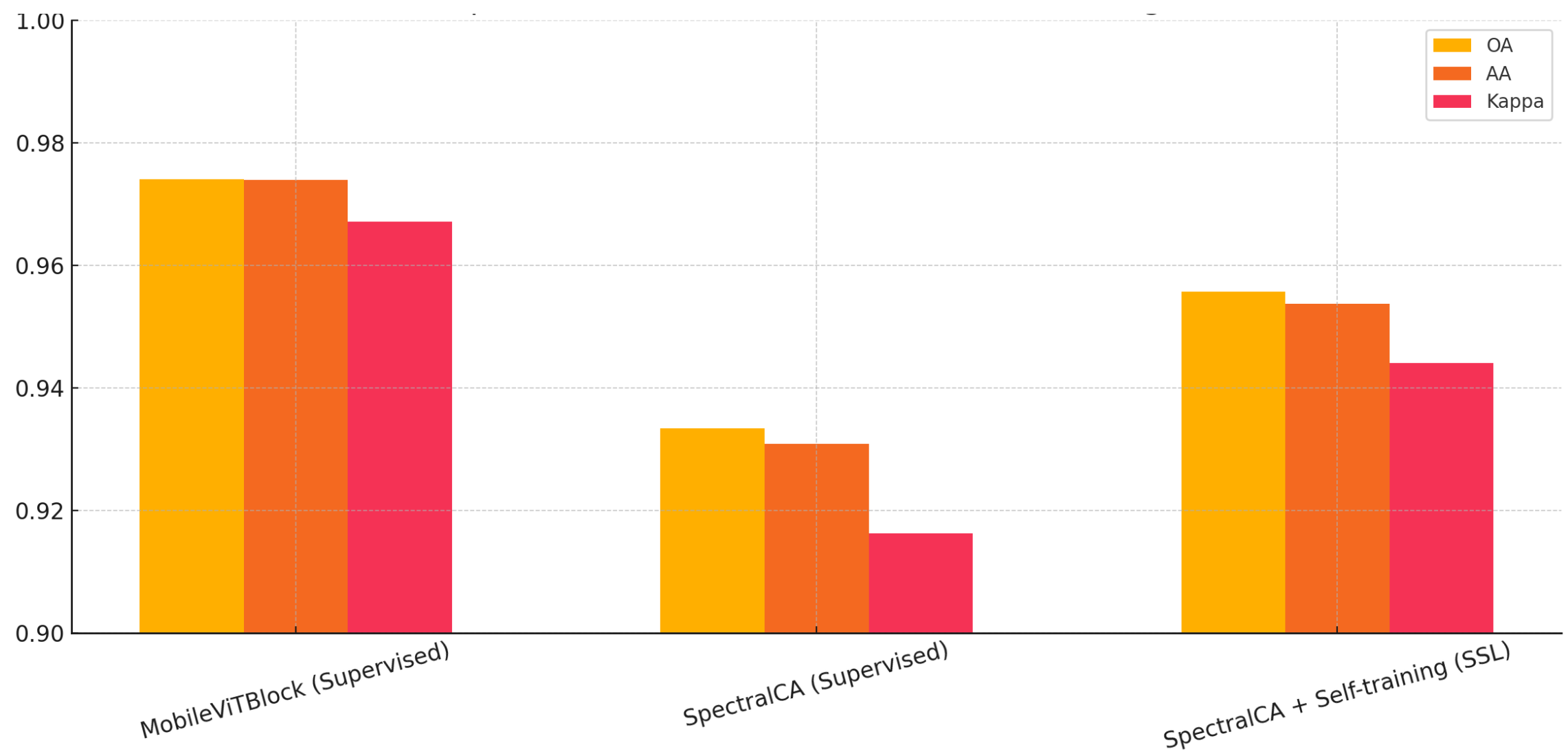

**Fig. 3.5** – Comparison of model accuracy

The plot in Fig. 3.5 shows that after semi-supervised learning, the accuracy of the model increased, approaching the initial version of MDvT.

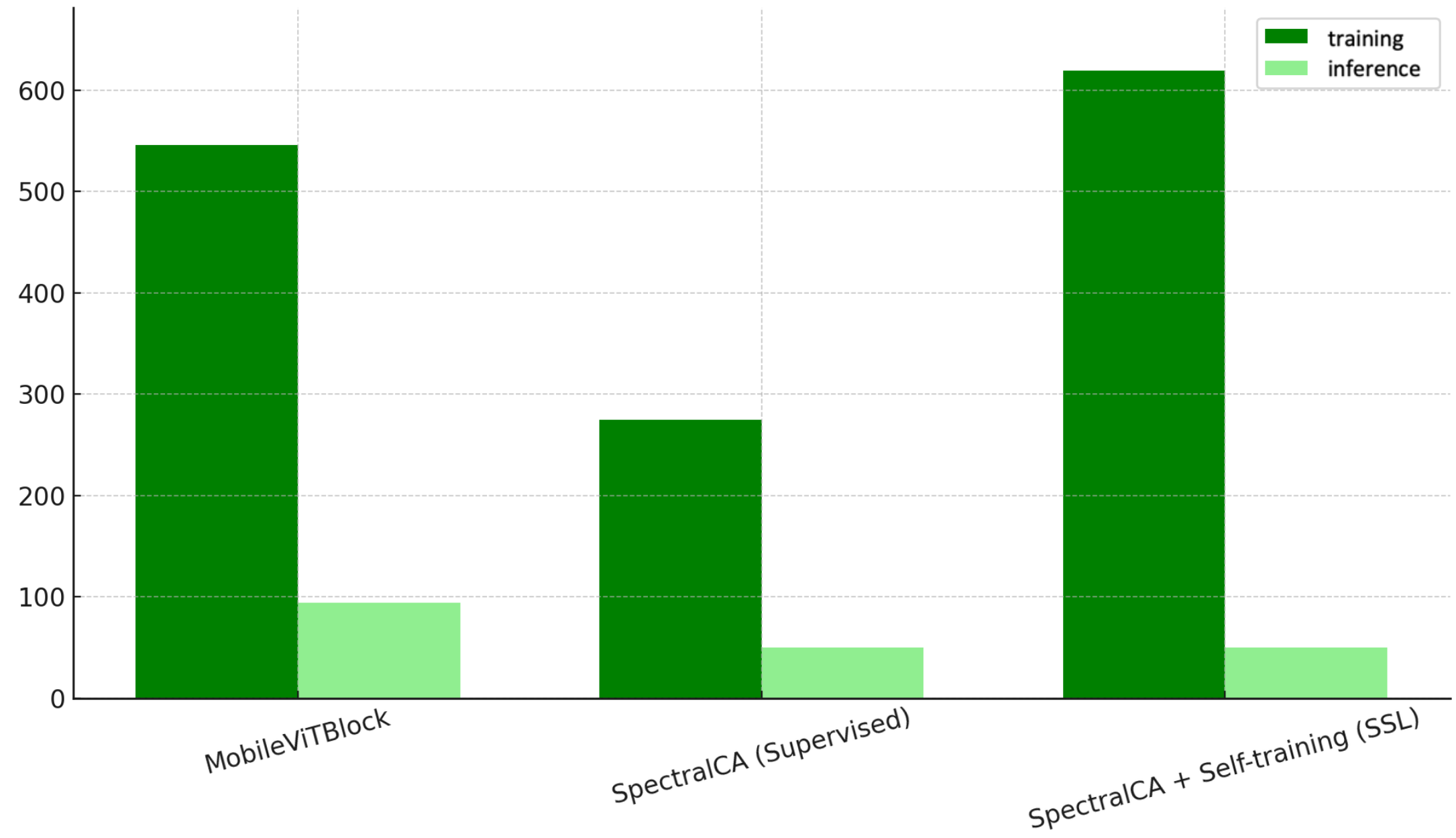

**Fig. 3.6** – Comparison of model operating times in seconds

Fig. 3.6 demonstrates the stable inference time of MDvT with the SpectralCA block. However, it is evident that although the inference time remained twice as fast as MDvT with MobileViTBlock, the training time increased due to the larger amount of data.

Table 3.7 contains more detailed characteristics of SpectralCA usage in supervised and semi-supervised training. It shows that the accuracy of the model in SSL increased by ~2%, reaching ~95%.

**Table 3.7** – SpectralCA metrics in supervised and semi-supervised training runs

| Setup | OA | AA | Kappa | Train time (sec) | Infer time (sec) | Param count (mil) |
|---|---|---|---|---|---|---|
| **SpectralCA (Supervised)** | 0.9334 | 0.9308 | 0.9162 | **274.38** | 50 | 6.628 |
| **SpectralCA + Self-training (SSL)** | **0.9557** | **0.9537** | **0.944** | 619.05 | 50 | 6.628 |

The experiments demonstrated that the use of self-training with proxy labeling improves model accuracy. Five thousand new images were labeled. The method allowed the model to be adapted to new image fragments that were not included in the initial training set and reduced dependence on fully annotated datasets. This approach is particularly relevant for practical application in the field, where the number of verified labels is limited and the need for accurate object recognition is critically high.

Table 3.8 shows a detailed comparison of MobileViTBlock and SpectralCA + Self-training. The accuracy of the proposed model is only ~2% lower than the initial architecture with MobileViTBlock, while maintaining twice as fast inference and 1.1 million fewer parameters.

**Table 3.8** – Finalized comparison between MobileViTBlock and SpectralCA with Self-training

| Setup | OA | AA | Kappa | Train time (sec) | Infer time (sec) | Param count (mil) |
|---|---|---|---|---|---|---|
| **MobileViTBlock** | **0.974** | **0.9739** | **0.9671** | **545.64** | 94 | 7.746 |
| **SpectralCA + Self-training (SSL)** | 0.9557 | 0.9537 | 0.944 | 619.05 | **50** | **6.628** |

Thus, the integration of semi-supervised learning with architecture that includes SpectralCA blocks is a promising direction for the development of hyperspectral classification systems for real-world tasks.

# CONCLUSION

A new hybrid neural network architecture for hyperspectral image classification was developed, based on a modification of the MDvT model with the introduction of a SpectralCA (Spectral Cross-Attention) block. The proposed module enables more effective modeling of the relationships between spectral and spatial features through a bidirectional cross-attention mechanism, which integrates information from both types of features in a symmetric and complementary way. This approach reduces the number of model parameters and halves both training and inference time compared to the baseline MobileViTBlock, while still maintaining high classification accuracy.

Experimental analysis showed that although the new architecture is slightly behind the baseline version in terms of absolute metric values (OA, AA, Kappa), it provides significant improvements in speed and resource efficiency. This is particularly important for deployment in resource-constrained environments, such as mobile devices and unmanned aerial vehicles (UAVs) operating in real-time field conditions. The reduced parameter count also facilitates adaptation of the model to new tasks and simplifies integration with other modules within larger remote sensing or environmental monitoring systems.

A notable advantage of the proposed architecture is its flexibility: the SpectralCA module can be integrated either as a full replacement for individual blocks in the transformer structure or as an additional component in the early layers of the model. This flexibility opens opportunities for further refinement and adaptation of the network to specific data types or application domains. In particular, one promising direction is the application of the model to UAV computer vision tasks, where the combination of real-time inference speed and hyperspectral feature discrimination could significantly enhance environmental perception, terrain classification, and autonomous decision-making.

A particularly important component of the implemented system was the integration of a semi-supervised learning method based on self-training with proxy labeling.

Thanks to the iterative expansion of the training set using highly probable pseudo-labels, it was possible to improve the accuracy of the model even in cases where the amount of manually annotated data was limited. This opens up prospects for using the model in real-world conditions where complete labeling is not available, and also lays the foundation for the future use of more advanced semi-supervised or active learning strategies.

In conclusion, the results of the study demonstrate the feasibility and effectiveness of using a modified MDvT hybrid architecture with a SpectralCA cross-attention block and the application of self-training in hyperspectral image classification tasks. The proposed model demonstrates strong potential for real-world use in environmental monitoring, agriculture, UAV-based navigation, security applications, and other domains where accuracy, efficiency, and adaptability to limited resources are critical.

## APPENDIX A SOURCE CODE OF THE SPECTRALCA MODULE

The SpectralCA module is implemented in Python, utilising the PyTorch framework. The proposed block leverages bi-directional cross-attention to enhance the modeling of spectral-spatial feature dependencies.

URL: https://github.com/BrovkoD/spectral-cross-attention.